\documentclass{svproc}

\usepackage{graphicx}
\usepackage{booktabs}
\usepackage{multirow}
\usepackage{marvosym}

\usepackage{url}

\begin{document}
\mainmatter              
\title{Amodal Segmentation for Laparoscopic Surgery Video Instruments}
\titlerunning{Amodal Surgery Instruments Segmentation}  
%
\author{Ruohua Shi\inst{1,2}$^{\star}$, Zhaochen Liu\inst{1,2,3}\thanks{These authors contributed equally.}, Lingyu Duan\inst{1,2}, Tingting Jiang\inst{1,2,4}\Letter}
\authorrunning{Shi et al.} 
%
%
\institute{National Engineering Research Center of Visual Technology, School of Computer Science, Peking University, Beijing, China\\ \and
State Key Laboratory of Multimedia Information Processing, School of Computer Science, Peking University\\ \and
AI Innovation Center, School of Computer Science, Peking University\\ \and 
National Biomedical Imaging Center, Peking University\\
\email{\{shiruohua,dreamerliu,lingyu,ttjiang\}@pku.edu.cn}
}

\maketitle              

\begin{abstract}
Segmentation of surgical instruments is crucial for enhancing surgeon performance and ensuring patient safety. Conventional techniques such as binary, semantic, and instance segmentation share a common drawback: they do not accommodate the parts of instruments obscured by tissues or other instruments. Precisely predicting the full extent of these occluded instruments can significantly improve laparoscopic surgeries by providing critical guidance during operations and assisting in the analysis of potential surgical errors, as well as serving educational purposes. In this paper, we introduce Amodal Segmentation to the realm of surgical instruments in the medical field. This technique identifies both the visible and occluded parts of an object. To achieve this, we introduce a new Amoal Instruments Segmentation (AIS) dataset, which was developed by reannotating each instrument with its complete mask, utilizing the 2017 MICCAI EndoVis Robotic Instrument Segmentation Challenge dataset. Additionally, we evaluate several leading amodal segmentation methods to establish a benchmark for this new dataset.

\end{abstract}

%

\section{Introduction}
As minimally invasive surgical robots advance, computer vision and machine learning-based assistive systems have become increasingly crucial in boosting surgeon performance and patient safety. However, numerous challenges arise in analyzing the data captured by surgical cameras. Surgical instrument segmentation serves as a critical and indispensable element in a range of computer-assisted interventions.

\begin{figure}[t]
\includegraphics[width=\textwidth]{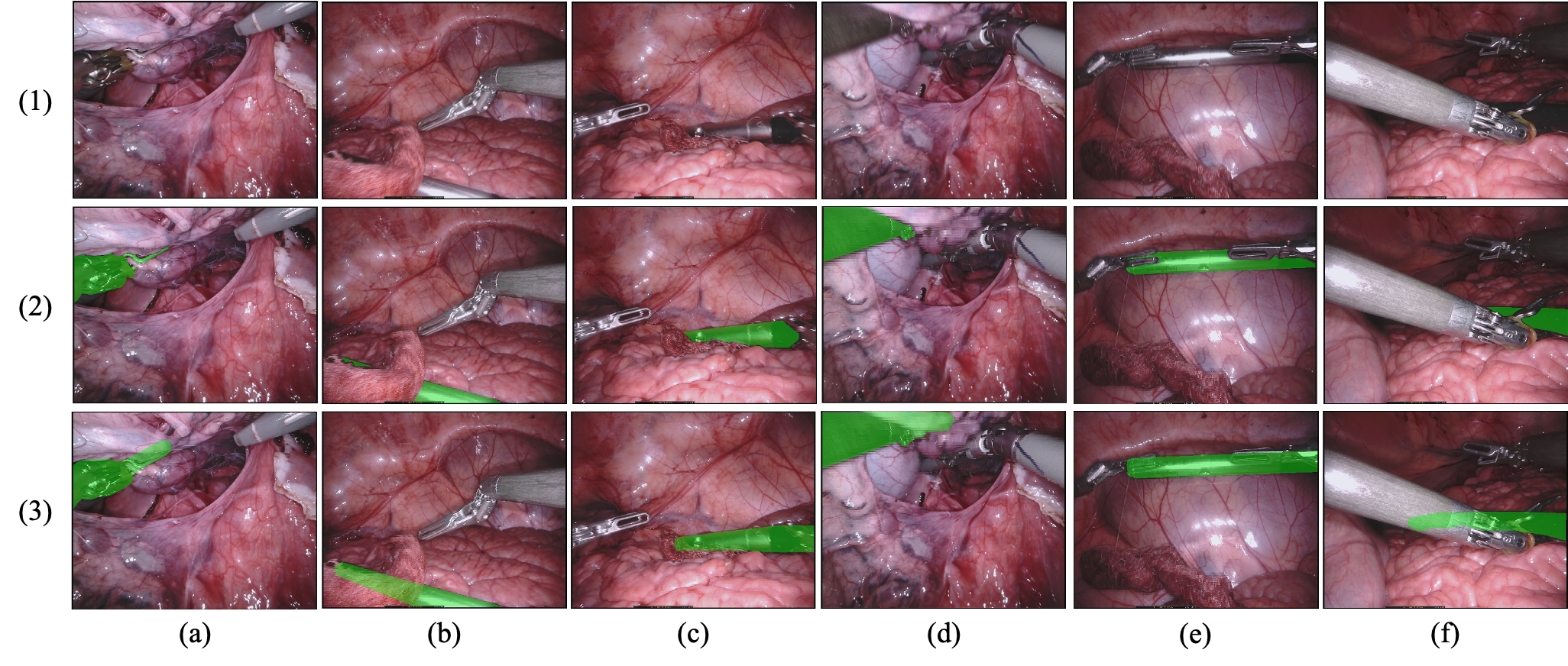}
\caption{Segmentation examples. (1) Frames from laparoscopic surgery; (2) Instance segmentation ground truth; (3) Segmentation masks with occluded parts.} 
\label{fig1}
\end{figure}

Recent research has made strides in addressing these challenges, with many researchers exploring solutions \cite{zhu2017semantic,TernausNet,MF-TAPNet,gonzalez2020isinet,sestini2023fun,yue2024surgicalsam}. Notably, a U-Net-based model \cite{TernausNet} clinched the top spot at the 2017 MICCAI EndoVis Robotic Instrument Segmentation Challenge \cite{2017challenge}. Subsequent studies, such as \cite{sestini2023fun}, have a generative-adversarial approach for unsupervised surgical tool segmentation of optical-flow images. Yet, these advancements predominantly focus on the visible portions of surgical instruments. Occlusions, whether between instruments and tissues or among the instruments themselves, can obscure critical data during clinical procedures, complicating the surgical process. For illustration, Fig.~\ref{fig1} shows the selected frames from laparoscopic surgery videos. The second and third rows distinctly display instrument masks with and without occlusions. It is evident that standard instance segmentation masks fail to retain the original shape of the instruments, thereby complicating the identification of instrument types from a single frame.

Therefore, accurately predicting occluded instruments can be beneficial in surgeries. Specifically, there are three primary applications: (a) During surgery, predicting the obscured parts of surgical instruments can assist surgeons by providing crucial visual cues about the position and orientation of the tools, ensuring the accuracy of their maneuvers. For example, if the occluded head of an instrument is accurately predicted, surgeons can better judge the alignment of the instrument relative to the target area. (b) Post-operatively, accurate reconstruction of the surgical process through video analysis allows medical professionals to assess whether the instruments were used correctly and if the procedures met the required standards. This includes determining the types of surgical instruments employed and evaluating the precision of the interactions with the patient, thereby assessing the potential impact on patient safety, including any adverse events. (c) For educational purposes, accurately tracking the key points and trajectories of surgical instruments allows instructors to dissect and elucidate optimal surgical techniques. It also highlights crucial visual indicators during procedural training.

Segmenting both the visible and occluded regions of each object instance simultaneously is referred to as \textbf{Amodal Segmentation}. This concept is a recent development that builds upon instance-aware segmentation. Numerous methodologies have been introduced to tackle this issue \cite{follmann2019learning,ke2021deep,tran2022aisformer,wang2020robust}. To the best of our knowledge, the recent amodal algorithms that achieved state-of-the-art performance is SAM-based~\cite{liu2024plug}, exploiting the powerful feature extraction capability provided by the large-scale foundation model~\cite{kirillov2023segment}. However, all these amodal segmentation approaches have predominantly been applied to natural images and have not been adapted for the medical field.

In this work, for the first time, we introduce the amodal segmentation to the area of laparoscopic surgery, and propose the first amodal dataset for surgical instruments, named \textbf{AIS}, which is based on the 2017 Robotic Instrument Segmentation Challenge dataset~\cite{2017challenge}. Unlike traditional datasets, AIS includes labels for the full mask of each instrument, covering both visible and occluded regions. Additionally, we evaluated several leading amodal segmentation methods to establish a benchmark for this innovative dataset.

In summary, our paper contains the following contributions: 

\begin{itemize}
    \item For the first time, amodal segmentation is applied to surgical instruments within the medical field. 
    \item A novel medical amodal segmentation dataset is developed specifically tailored for this task.
    \item A benchmark is introduced to evaluate the accuracy of predicting instruments, including their occluded parts, using our new dataset.
\end{itemize}

\section{Related Work}
\subsection{Surgery Video Instrument Segmentation} Various advancements have been made in semantic segmentation of surgical surgery, incorporating various techniques and methodologies, mainly including combining segmentation networks with attention mechanisms~\cite{ni2020attention,shen2023branch} and exploring some synthetic data for semi-supervised training~\cite{sestini2023fun,garcia2021image}. Ni et al. propose an attention-guided lightweight network, utilizing depth-wise separable convolution as the basic unit to reduce computational costs, thereby performing surgical instrument segmentation in real-time~\cite{ni2020attention}. Shen et al. employ a lightweight encoder and branch aggregation attention mechanism to remove noise caused by reflection, and water mist to improve segmentation accuracy and achieve a lightweight model~\cite{shen2023branch}. Sestini et al. designed a fully unsupervised method for the segmentation of binary surgical instruments relying only on implicit motion information and a priori knowledge of the shape of the instrument~\cite{sestini2023fun}. Luis et al. facilitate this task by generating large amounts of trainable data by synthesizing surgical instruments with real surgical backgrounds~\cite{garcia2021image}. However, previous approaches mainly focus on real-time or semi-supervised, unsupervised learning, and there has been no research on interactive segmentation. Moreover, their methods cannot distinguish the occluded instruments. Therefore, we explored the segmentation of both visible and occluded instruments in surgery.

\subsection{Amodal Segmentation}
Amodal segmentation task was proposed in 2016 to predict the complete shape of target object including both the visible parts and the occluded parts~\cite{li2016amodal}. Possessing great significance to our seeking visual intelligence, amodal segmentation arouses increasing attention in the academic community. Besides direct methods~\cite{li2016amodal,zhu2017semantic,qi2019amodal}, a series of elaborate approaches have been designed with diverse concepts involved to achieve better performance, such as depth relationship~\cite{zhang2019learning}, region correlation~\cite{follmann2019learning,ke2021deep,tran2022aisformer}, shape priors~\cite{xiao2021amodal,li20222d,gao2023coarse}, and compositional models~\cite{wang2020robust}. Recently, a SAM-based approach~\cite{liu2024plug} achieved state-of-the-art performance, well exploiting the mighty feature extraction capability provided by the large-scale foundation model~\cite{kirillov2023segment}. Noticing the labor-intensive and error-prone challenges in the annotation of amodal masks, researchers also propose many weakly supervised approaches for amodal segmentation using only box-level supervision or self supervision~\cite{zhan2020self,nguyen2021weakly,kortylewski2020compositional,kortylewski2021compositional,sun2022amodal,liu2024blade}. 

Based on the advances of amodal segmentation algorithms, researchers develop various applications. For example, utilizing amodal segmentation methods, intelligence systems like autonomous driving and robotic grasping can enhance the safety and the reliability~\cite{qi2019amodal,breitenstein2022amodal,wada2018instance,wada2019joint,inagaki2019detecting}, while many new implementation paths emerge in the fields of diminished reality~(DR) and novel view synthesis~\cite{gkitsas2021panodr,li20222d,pintore2022instant}. Though numerous related work is delivered, the potential of amodal segmentation in the medical field has never been explored. With our newly released dataset, we select several typical segmentation methods for experiments, revealing the notable value in medical applications.

\section{Dataset}

We introduce the AIS dataset by re-annotating the 2017 Robotic Instrument Segmentation Challenge dataset~\cite{2017challenge}. The dataset consists of 10 videos, each with 300 frames. It covers different abdominal porcine procedures recorded using the da Vinci Xi systems. In this paper, we relabeled the 3000 frames at a resolution of 1024$\times$1280. The annotation includes semantic instance-level amodal segmentation and each object is manually assigned a class label. There are a total of 7084 objects. The dataset is split into training and test sets based on the splitting rule of the 2017 Robotic Instrument Segmentation Challenge: take the first 225 frames of 8 sequences as training data and keep the last 75 frames of those 8 sequences as test data. 2 of the full 300 frame sequences were kept as test sequences.

\subsection{Dataset Acquisition}
To annotate the dataset, we employed a widely-used public annotation tool named \emph{labelme} \cite{labelme}. This tool allows for the interactive labeling of surgical instruments on a frame-by-frame basis, enabling manual annotations directly within the software interface.

\begin{figure}[t]
    \includegraphics[width=\textwidth]{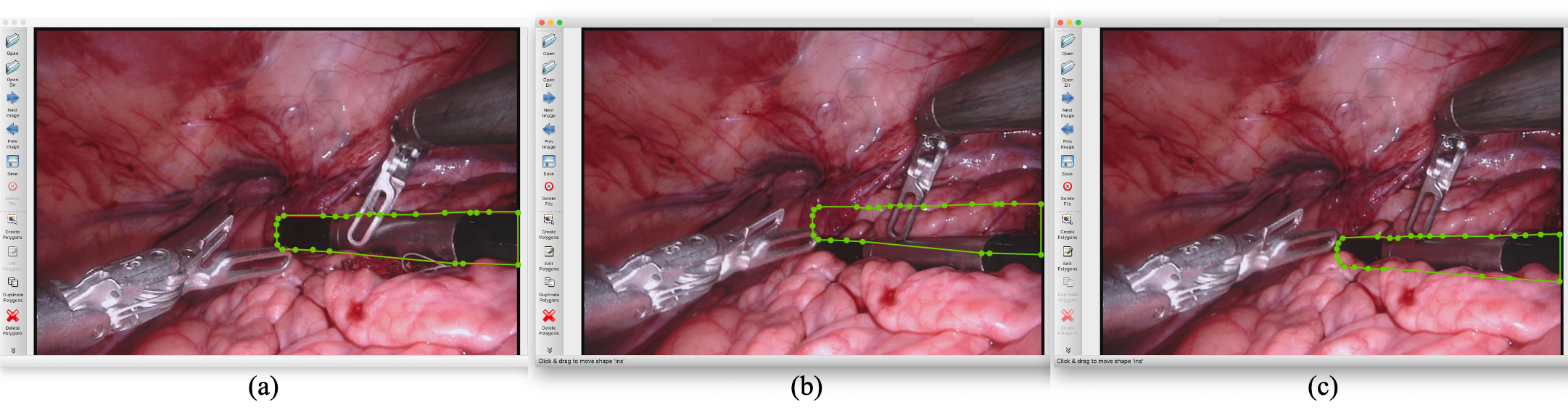}
    \caption{Illustration of annotation process. (a) displays a polygon annotation formed by various key points and line segments. Once the labeling of this frame is complete, the annotation can be carried forward to the next frame for further labeling, as depicted in (b). By adjusting the positions of the key points within the polygon, the final annotation of the instrument in (b) is presented in (c).} 
    \label{fig2}
\end{figure}

Consider a practical example of our annotation methodology as illustrated in Fig.\ref{fig2}(a), the middle instrument is annotated with a polygon that includes several key points and line segments. In this interface, annotations can be refined by adjusting the positions of key points (illustrated as green points). After completing the annotation for one frame, it can be carried over to the next frame, as depicted in Fig.~\ref{fig2}(b). Given that the movement of surgical instruments between consecutive frames is typically minimal, the annotation for the current frame can often be achieved by simply repositioning the key points from the previous frame to align with the instrument's shape. This process completes the annotation for the middle instrument in the current frame, with the final result displayed in Fig.~\ref{fig2}(c). By employing this annotation strategy, we attempt to reconstruct the original form of the instruments as accurately as possible, even for occluded parts. Our guiding principle is to ensure that the prediction of the occluded sections adheres closely to the instrument’s actual shape and follows its natural motion trajectory.

\begin{table}[b]
    \centering
    \caption{Amodal segmentation dataset statistics.}
    \label{tab1}
    \begin{tabular}{l|c|c|c|c}
    \midrule
    $\ $\textbf{Dataset} &  $\ \ $\textbf{COCOA}$\ \ $ & $\ \ $\textbf{COCOA-cls}$\ \ $ & \textbf{D2S}  & \textbf{Ours}\\
    \midrule
    $\ $Image/Video &  Image & Image & Image & Video\\
    \midrule
    $\ $Resolution & 275K pix & 275K pix & 3M pix & 1M pix\\
               & -   & -  & $\ \ $1440$\times$1920$\ \ $ & $\ \ $1024$\times$1280$\ \ $\\
    \midrule
    $\ $\# of images  & 5073 & 3499 & 3499 & 3000\\
    $\ $\# of instances & 46314  & 10562 & 28720 & 7084\\
    $\ $\# of occluded instances$\ $ & 28106 & 5175 & 16337  & 1455\\
    \midrule
    $\ $Avg. occlusion rate & 18.8\% & 10.7\% & 15.0\% & 20.54\%\\
    \midrule
    \end{tabular}
\end{table}

\subsection{Dataset Analysis}


We compare the proposed AIS dataset with other prominent amodal segmentation datasets from real-world scenes, such as COCOA\cite{cocodataset}, COCOA-cls\cite{cocodataset,follmann2019learning}, and D2S \cite{follmann2019learning,D2S}. Specifically, we examine the differences in resolution, the number of instances included, and the rate of occlusion across these datasets. As shown in Table~\ref{tab1}, our dataset exhibits a higher average occlusion rate and represents the first amodal segmentation dataset specifically developed for the medical field.

\section{Benchmarks}
\subsection{Selected Methods}
Our newly released dataset brings novel medical application scenarios for amodal segmentation. To better reveal the capabilities of existing approaches, we select several typical methods and benchmark their performance on this dataset, which are SAM~\cite{kirillov2023segment}, AISFormer~\cite{tran2022aisformer}, C2F-Seg~\cite{gao2023coarse}, and PLUG~\cite{liu2024plug}. 

SAM, the segment anything model, is the most influential foundation model for general segmentation task, which is trained on billions of object masks thus possessing strong abilities on diverse downstream tasks. The other three methods are specifically designed for amodal segmentation. PLUG is the state-of-the-art approach now, which is based on SAM and introduces the parallel LoRA structure and the uncertainty guidance. AISFormer and C2F-Seg are other two most recent methods. AISFormer first injects the transformer backbone into the amodal segmentation task, while C2F-Seg introduces the vector-quantization model and the coarse-to-fine framework.

\subsection{Evaluation Metrics}
As related work, we choose intersection-over-union (IoU) as the evaluation metric. The IoU of a predicted mask equals the ratio of the intersecting area with corresponding ground truth mask to the area of their union. In order to better reflect the performance, we calculate the average IoU on each sub-testset~(from 1 to 10) and the mean IoU of all sub-testsets.

\subsection{Experimental Setup}

\begin{figure}[t]
  \centering
   \includegraphics[width=\linewidth]{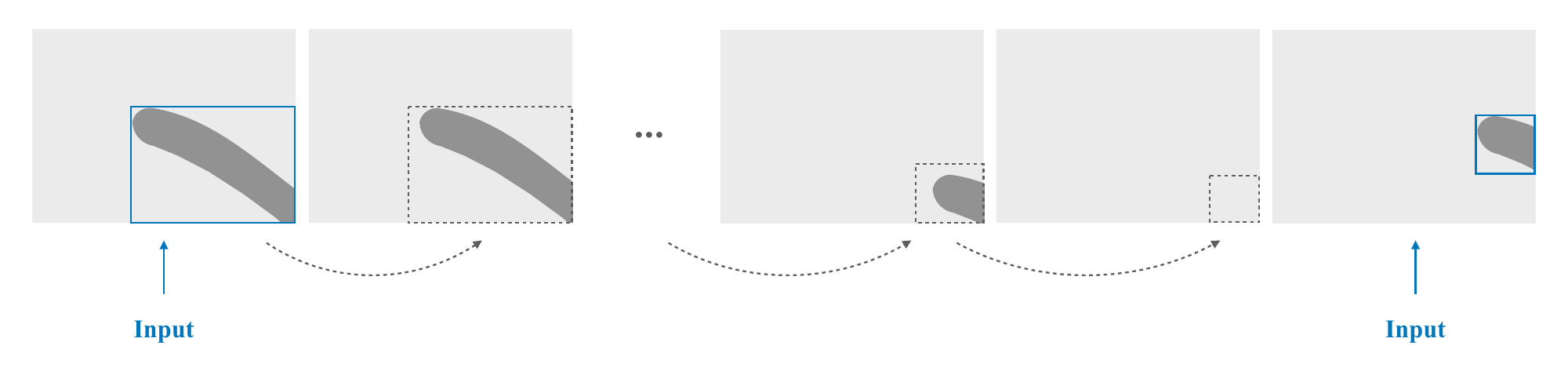}
   \caption{\textbf{The input mode of bounding boxes.} In order to reduce the number of manual inputs, we use the bounding box of the predicted mask in the previous frame as the input bounding box of the next frame when possible.}
   \label{fig:bbox}
\end{figure}

For SAM, we adopt the largest ViT-H version pretrained model to achieve its best performance. For the other three methods, we conduct training on our proposed dataset. The settings and parameters adopted during the training process are consistent with the original papers.

It is worth explaining that the input mode of bounding boxes. Since the models require clear regions of interest, we input bounding boxes as guidance in all these methods. SAM, AISFormer, and PLUG directly receive input bounding boxes, while C2F-Seg uses the predictions of SAM as input thus indirectly receiving the bounding boxes~(C2F-Seg generates amodal mask based on the input visible mask. For fairness, we adopt the output of SAM as the input visible mask of C2F-Seg.). Considering practical needs, we hope to reduce the number of manual inputs as much as possible for the convenience of users. Actually, we notice that the difference between two adjacent frames of a video is very small, and we can approximately assume that the bounding boxes in the two frames are almost the same. So as shown in Fig.~\ref{fig:bbox}, we first input the bounding box of the first frame, and then use the bounding box of the predicted mask of the previous frame for each subsequent frame until a certain frame without the corresponding bounding box in the previous frame occurs, we then manually input the bounding box again.

\section{Results}

\subsection{Performance comparison}
As shown in Table~\ref{tab:comparison}, all these segmentation methods can basically predict the amodal masks of surgery instruments quite well. The pretrained SAM without any fine-tuning can already reach practical performance, achieving 85.94 on mean IoU. Specifically designed, the other three methods can provide more accurate predictions. AISFormer and C2F-Seg attain 86.65/88.17 on mean IoU, which are 0.71/2.23 higher than SAM. The state-of-the-art method PLUG maintains the leading position and reaches 89.25 on the mean IoU, which beats SAM by 3.31. It can be observed that the performance is close on some relatively simple sub-testsets, such as testset-4, while on some complex sub-testsets like testset-1, the advantage of tailored amodal segmentation methods especially PLUG is evident.

\begin{table}[t]
\caption{The performance comparison of selected segmentation methods on the surgery dataset. Numbers 1-10 represent ten sub-testsets.}
\label{tab:comparison}
\resizebox{\linewidth}{!}{%
\begin{tabular}{c|cccccccccc|c}
\midrule
Methods   & 1    & 2    & 3    & 4    & 5    & 6    & 7    & 8    & 9    & 10   & $\ $Mean$\ $           \\ \midrule
SAM~\cite{kirillov2023segment}       & 75.1 & 80.5 & 91.5 & 93.1 & 81.1 & 85.8 & 83.7 & 91.2 & 85.9 & 91.5 & 85.94 \\ 
$\ $AISFormer~\cite{tran2022aisformer}$\ $ & $\ $74.6$\ $ & $\ $82.6$\ $ & $\ $91.0$\ $ & $\ $93.6$\ $ & $\ $83.0$\ $ & $\ $83.9$\ $ & $\ $88.6$\ $ & $\ $93.0$\ $ & $\ $85.2$\ $ & $\ $91.0$\ $ & 86.65 \\ 
C2F-Seg~\cite{gao2023coarse}   & 75.8 & 85.6 & 92.6 & 93.5 & 85.0 & 86.1 & 89.1 & 94.2 & 87.0 & 92.8 & 88.17 \\ 
PLUG~\cite{liu2024plug}      & 79.8 & 86.2 & 93.2 & 93.6 & 86.1 & 87.1 & 90.6 & 94.4 & 88.4 & 93.1 & 89.25 \\ \midrule
\end{tabular}%
}
\end{table}

To intuitively display the performance of these methods, we choose one frame from each sub-testset, totaling ten frames. The ten sets of qualitative results are shown in Fig.~\ref{fig:res}. Relatively speaking, though the predictions of other methods are acceptable, PLUG can determine more precise and complete shapes. For the rod-shaped instrument in the 1st row, the prediction of PLUG is clear to identify showing smooth boundary and covering more area. For the semi-inserted instrument in the 8th row, C2F-Seg and PLUG can segment the needle head, while the prediction of PLUG is thinner and closer to reality.


\subsection{Hard Case}
\begin{figure}[h]
  \centering
   \includegraphics[width=\linewidth]{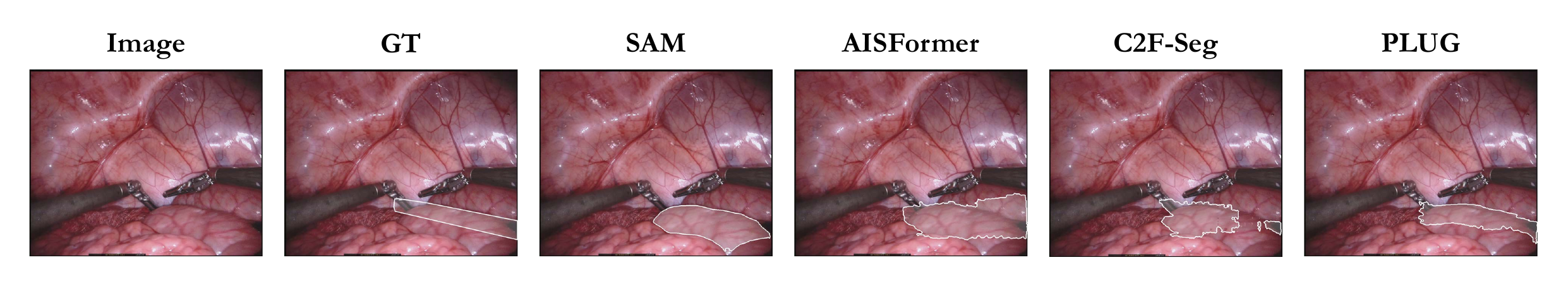}
   \caption{\textbf{An example of the excessively occluded case.} The instrument is almost completely occluded in this frame.}
   \label{fig:hard}
\end{figure}

\begin{figure}[t]
  \centering
   \includegraphics[width=\linewidth]{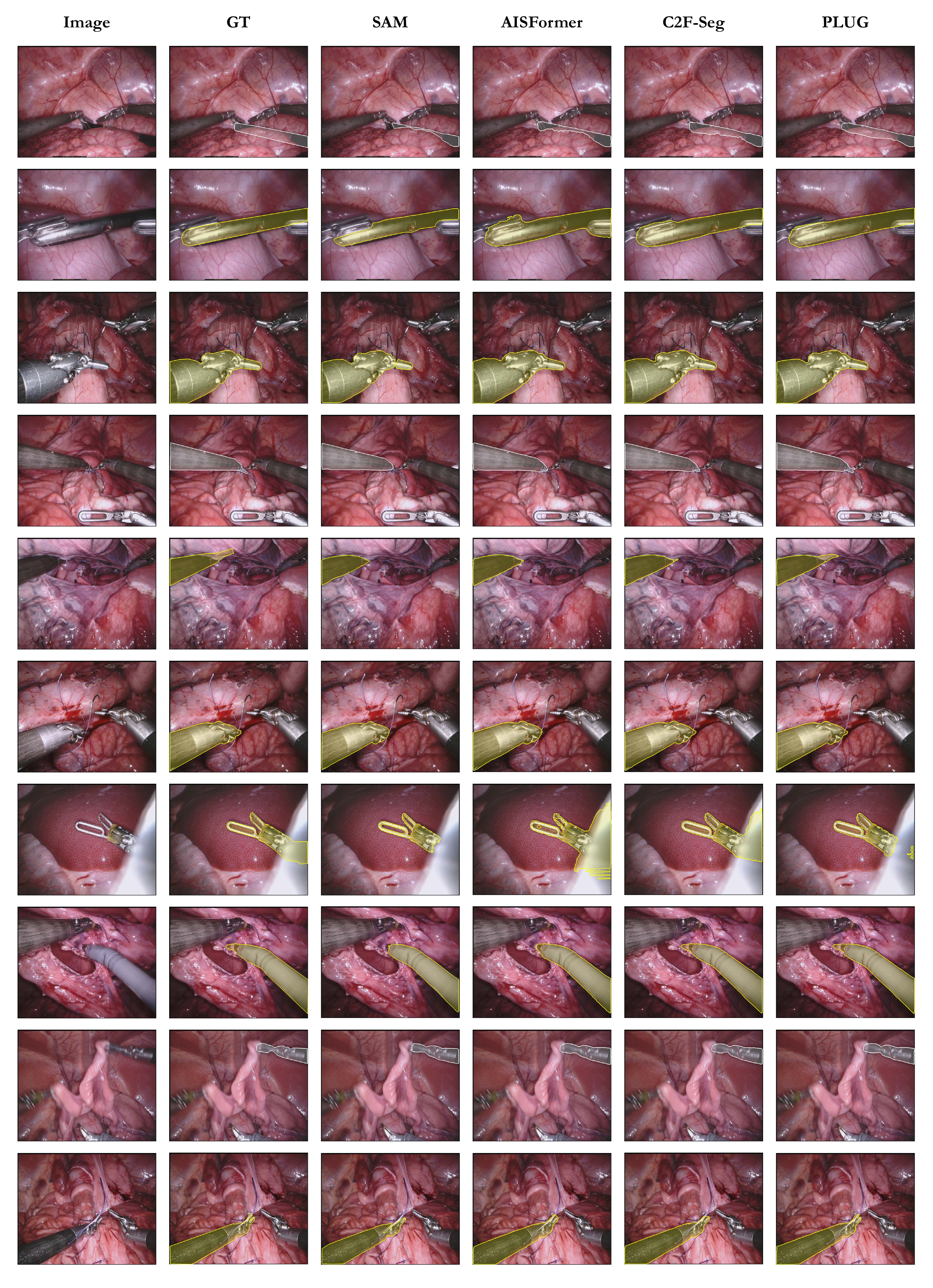}
   \caption{\textbf{Qualitative results.} The qualitative comparison of predicted amodal masks from SAM, AISFormer, C2F-Seg and PLUG. These ten rows are chosen from the first to tenth sub-testset in order from top to bottom.}
   \label{fig:res}
\end{figure}

In the experiment, we find that some excessively occluded instruments~(as shown in Fig.~\ref{fig:hard}) in a fraction of frames are a common challenge for these methods. These instruments are almost completely occluded, making it difficult to obtain effective visible mask or other information, thus bringing confusion to methods inspired by extracted clues within the frame. In the example shown in Fig.~\ref{fig:hard}, even the state-of-the-art PLUG approach predicts a quite tortuous boundary. To tackle the hard case, a pertinent method designed for video input that integrates the context of the previous and subsequent frames may be needed. How to implement such a specific method? We leave this issue to future work.

%
%
\clearpage
\bibliographystyle{splncs03}
\bibliography{AIS}

\end{document}